\documentclass[letterpaper, 10 pt, conference]{ieeeconf}  

\IEEEoverridecommandlockouts                              

\usepackage{hyperref}
\usepackage{float}
\usepackage{graphicx}

\title{\LARGE \bf Disambiguating Affective Stimulus Associations for Robot Perception and Dialogue}
\author{Henrique Siqueira*, Alexander Sutherland*, Pablo Barros, Mattias Kerzel, Sven Magg, Stefan Wermter\\
\\\\\large{\textbf{Pre-print version of \cite{8625012}}}
	\thanks{Knowledge Technology, Department of Informatics, University of Hamburg, Germany, \{{\tt\small siqueira, sutherland, barros, kerzel, magg, wermter\}@informatik.uni-hamburg.de}} \thanks{* indicates equal contribution}}

\begin{document}

\maketitle
\thispagestyle{empty}
\pagestyle{empty}

\begin{abstract}
Effectively recognising and applying emotions to interactions is a highly desirable trait for social robots. Implicitly understanding how subjects experience different kinds of actions and objects in the world is crucial for natural HRI interactions, with the possibility to perform positive actions and avoid negative actions. In this paper, we utilize the NICO robot's appearance and capabilities to give the NICO the ability to model a coherent affective association between a perceived auditory stimulus and a temporally asynchronous emotion expression. This is done by combining evaluations of emotional valence from vision and language. NICO uses this information to make decisions about when to extend conversations in order to accrue more affective information if the representation of the association is not coherent. Our primary contribution is providing a NICO robot with the ability to learn the affective associations between a perceived auditory stimulus and an emotional expression. NICO is able to do this for both individual subjects and specific stimuli, with the aid of an emotion-driven dialogue system that rectifies emotional expression incoherences. The robot is then able to use this information to determine a subject's enjoyment of perceived auditory stimuli in a real HRI scenario.
\end{abstract}

\section{INTRODUCTION}    
Emotion recognition, the task of identifying external features associated with internal emotional processes, has recently received significant attention from the research community. The ability to recognise and act on emotions has been identified as a core aspect of affective user-centric computing \cite{picard2003affective}. However, research attempting to actively apply recognised emotions is still in its formative stages, with recognised emotions rarely being applied to solving practical tasks. Instead, researchers have focused heavily on the process of recognising emotions in available datasets. 

Emotion incongruence is a phenomenon that occurs in multimodal input when different modalities indicate different emotions, such as an individual having a sad face but laughing leading to a conflicting understanding of an individual's emotional state. Exactly how humans handle this mismatch of multimodal information to reach a final conclusion is an ongoing field of research \cite{provost2015umeme,aguadoeffects2018}. Regardless, humans are often able to solve and understand these conflicting displays of emotions. Embodied agents, such as robots, will lack much of the context and dynamic situation handling that humans make use of to solve misunderstandings \cite{muller2011incongruence}. This lack of a holistic context is likely to become an ever more problematic element of HRI as we continue to attempt to integrate robots into different facets of human life.

\begin{figure}[t]
	\centering
	\includegraphics[width=.48\textwidth]{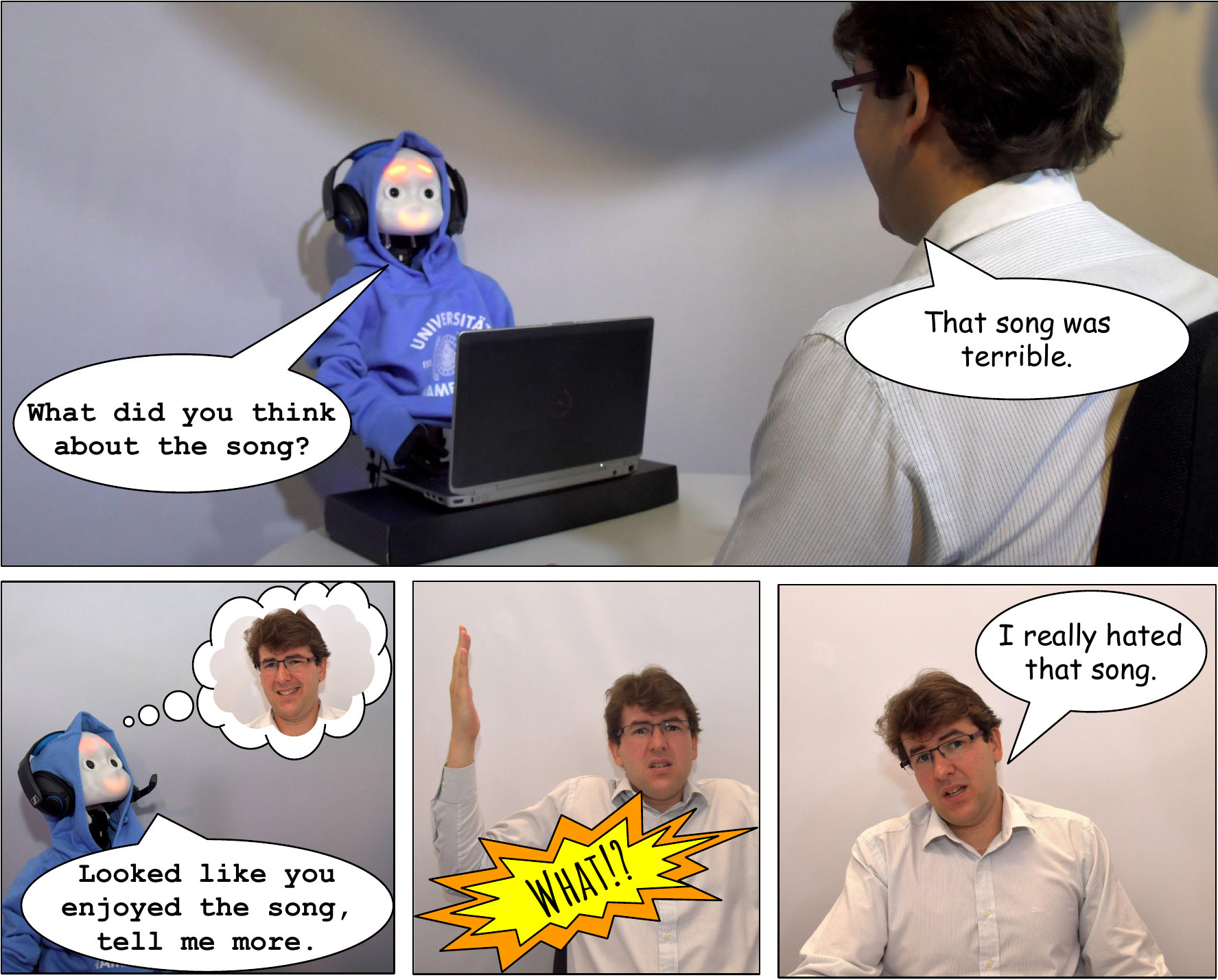}
	\caption{Illustration of our HRI experiment with the NICO robot under a mismatching condition where NICO initially perceived a positive visual affective response, but a negative response from speech. This situation leads NICO to proceed the dialogue in order to accrue affective information about the auditory stimulus.}
	\label{fig:experimentalsetup}
\end{figure}

As such, multimodal methods of learning what stimuli are experienced positively or negatively by users, in an implicit manner to avoid frustrating interactions, are required. Furthermore, practical methods of identifying and solving detected incoherences in multimodal perception need to be developed, as robots are likely to be exposed to contradictory expressions relating to various actions and objects. For these, the robot would ideally want to associate only positive or negative labels for the sake of simplified decision making. Our approach addresses both preferred stimuli learning from reactions and emotional perception coherency with a robot.

In this paper, we present an HRI scenario wherein a NICO robot, seen in Figure \ref{fig:experimentalsetup}, actively tracks an affective association between recognised emotional expressions and a triggering auditory stimulus, trying to determine if a particular stimulus was positively or negatively received. As misunderstandings between the visual and language modalities can occur, dialogue is used as a tool to acquire additional affective information about the association, with the ultimate goal of achieving a unified position on the positivity or negativity of the association.

\section{RELATED WORK}
Modelling perceived emotions in larger and more continuous contexts is a novel area of research. Barros et al. \cite{barros2017organizing} have presented initial work towards modelling perceived emotions in a dynamic context. In their work, the authors were able to show the emergence of different emotional concepts using Grow When Required (GWR) Networks \cite{marsland2002gwr}. The authors note that although an associative relation could be created, the conflict between modalities is not inherently handled by the system. Our approach opts for a simpler affective representation in order to focus more on the dynamics of general perception toward certain stimuli and the differences between individuals. We also present an initial step towards solving incoherent input in a manual fashion.

Some of the prevalent research that actively applies emotions to HRI scenarios can be described as falling into two application types: emotional colouring and emotion modulated behaviour. Emotional colouring approaches \cite{acosta2011achieving,zhou2017emotional} tend to focus on how to reciprocate recognised emotions, often with the intention of invoking empathy through mechanisms such as mirror neurons.

Acosta et al. \cite{acosta2011achieving} use learned patterns in emotional responses to change the prosody of audio responses, with the goal of achieving improved rapport with a user via an audible dialogue system called Gracie. Gracie's modulation of audio was based on the responses given by a student counselor to students throughout an interaction. The features of the counselor's responses to the students through the interactions were applied to Gracie's response system, allowing Gracie to mimic the response method. An evaluation showed that users preferred the emotion-specific modulation over a neutral or randomly varying version of the system. 
    
In the work of Zhou et al. \cite{zhou2017emotional}, emotions have been successfully applied to the generation of responses in a chat-bot scenario. By using a sequence to sequence encoder-decoder architecture, the authors were able to generate coherent and emotionally relevant responses in a chatbot scenario. The unique approach the authors used to achieve this was through the incorporation of an emotion category embedding as well as internal and external emotional memories that capture emotion dynamics during the decoding process of generating responses using a recurrent neural network. Word-by-word responses are generated with the generated words being influenced by the internal memory and the decision between emotional or non-emotional output words being influenced by the external memory.
 
While all of these colouring approaches are valuable contributions to enhancing HRI, they currently take a passive approach toward the role emotions in interactions. In these cases, emotion serves as a method to enhance an experience with a robot but does not directly contribute to a system's decision making procedure. Most emotion colouring scenarios could continue without a colouring factor and the scenario would still function, albeit in a less enjoyable manner. 

In contrast to emotional colouring approaches, some selected approaches choose to more actively applying recognised emotions to decision making in order to achieve some desired task. Kuhnlez et al. \cite{gonsior2011improving,kuhnlenz2013increasing} have applied recognised emotions to a Flobi robot's decision-making process, attempting to use recognised emotions to inspire a level of both empathy \textit{and} altruism in users. In doing so, users were more likely to assist the robot with a monotonous task for a longer period of time than when the robot did not apply this strategy. Still, this method relies on attempting to influence and manipulate the state of the human. In our approach, we seek to actively determine user-preferred user stimuli based on expressed emotions and to rectify conflicting expressions to stimuli through dialogue initiated and steered by recognised emotions.    

\section{AFFECTIVE ASSOCIATION MODELLING}
\begin{figure}[h]
	\centering
	\includegraphics[width=.49\textwidth]{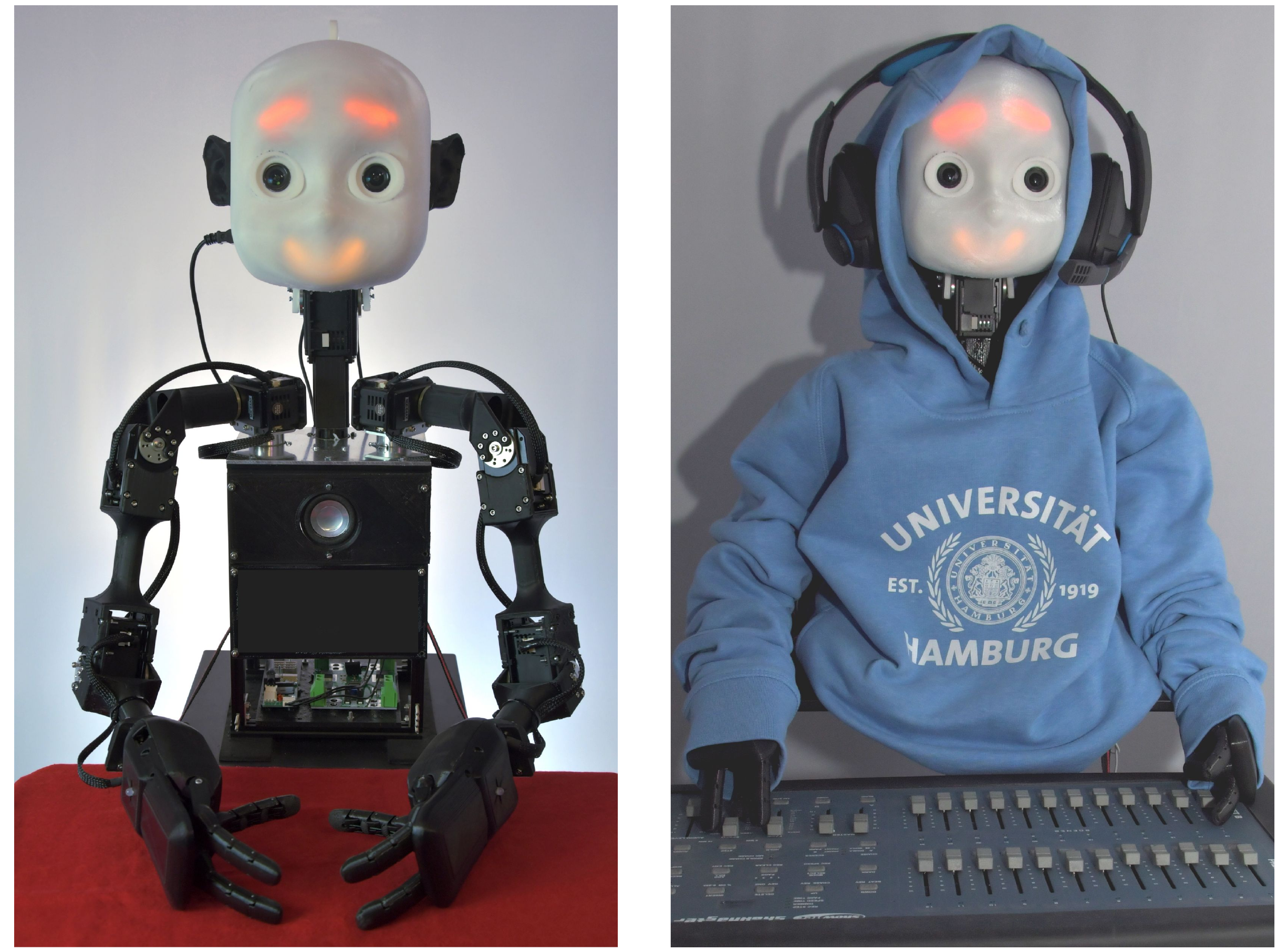}
	\caption{The Neuro-Inspired COmpanion robot. On the right, NICO is shown as a young DJ to better fit the HRI scenario.}
	\label{fig:nico}
\end{figure}

The experiments are realized using the humanoid robot platform NICO, the Neuro Inspired COmpanion. NICO, as seen in Figure \ref{fig:nico}, is a child-sized humanoid developed by the  Knowledge Technology Group of the University of Hamburg \cite{kerzel2017nico} as an open and highly customizable platform for research on neurobotic models and human-robot interaction. NICO's anthropomorphic design is modelled after a child with an age of three to four years. NICO is about one meter tall when standing up, it's head and arm proportions are matching for its size. For the reported experiment, we use the functionality of the upper body of NICO that was mounted on a fixed stand positioned behind a table to create the impression of being seated. NICO's upper body has two robotic arms based on human anatomy and range of motion and a head in the form of an abstracted child-like face with an emotion display and two cameras. 

NICO's arms have six degrees of freedom (DoFs) which are distributed as follows to enable human-like arm motions: three DoFs form a tight cluster in each shoulder, emulating the motion range of a shoulder ball joint, one DoF articulates the elbow, and two DoFs enable wrist rotation and flexation. NICO's end effectors are three-fingered HR4D tendon-operated hands from SeedRobotics\footnote{http://www.seedrobotics.com} which have the size of a child's hand. The shape of NICO's head is adapted from the iCub robot \cite{metta2008icub}. It has a symmetrical and abstracted child-like appearance that aims to enable intuitive human-robot interaction while avoiding the uncanny-valley effect. Behind the surface of the head, in the eyebrow and mouth area, a programmable LED display is placed, that can display basic emotions in the form of stylized facial expressions \cite{churamani2017teaching,churamani2018learning}. The head features two 2 Megapixel sensors with a 70-degree field of vision. The Head can perform pan and tilt motions. NICO's arms and head are articulated with Dynamixel servomotors and controlled via PyPot\footnote{https://github.com/poppy-project/pypot} and the open NICO API\footnote{www.inf.uni-hamburg.de/en/inst/ab/wtm/research/neurobotics/nico.html}.
 
Some effort has also gone into enhancing NICO's childlike qualities. As seen in Figure \ref{fig:operationcycle}, the NICO has been dressed to further enhance its childlike and anthropomorphic appearance. Making these qualities seem plausible is vital to creating a ``character'' that is supposedly interested in music and requires help. The overall appearance of the NICO has been matched with a childlike voice provided by Amazon Polly service\footnote{NICO uses the voice of Justin: https://aws.amazon.com/polly} to further enhance the impression of the NICO for participants during the interactions. NICO takes on the role of a DJ, having been provided with headphones and a laptop to further give the impression of a character that is playing and enjoys music. Subtle hand and head movements were made to enhance the perceived animacy of the NICO by making it a more animated interaction partner.

Our aim is to provide NICO with an effective mechanism for modelling affective associations between stimuli and emotional responses that takes into consideration mismatching emotional expressions from different modalities. Our approach consists of two affective perception modules to recognize emotional expressions based on facial expressions and cues from language, and a decision-making module to associate the perceived emotional response to a stimulus and a person. NICO plays a key role in this interaction as an anchor for subjects to interact with through dialogue and express emotions towards.


\subsection{Affective Visual Perception Module}
\begin{figure*}[t]
	\centering
	\includegraphics[width=.865\textwidth]{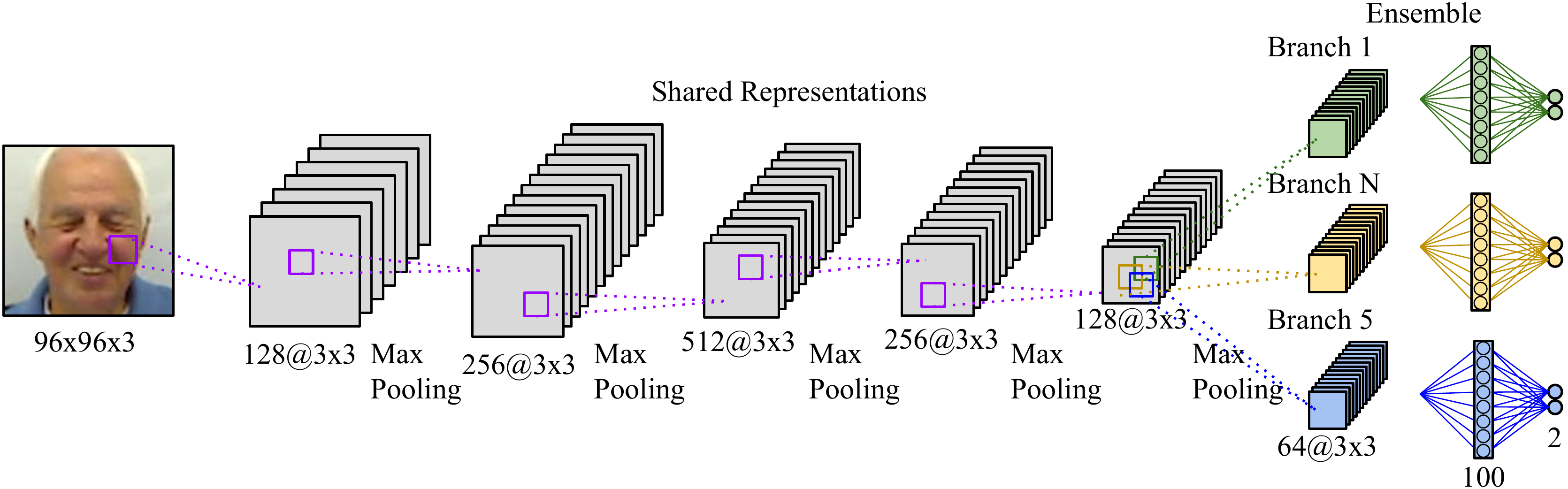}
	\caption{An ensemble with shared representations based on convolutional networks for arousal and valence recognition from a facial expression.}
	\label{fig:vision}
\end{figure*}

To recognize emotion in terms of arousal and valence from facial expressions, we adopted an ensemble with shared representations based on convolutional networks proposed by Siqueira et al. \cite{siqueira2018cnn} shown in Figure \ref{fig:vision}. Our architecture consists of a single branch with five convolutional layers (filter size of 3 x 3 with a stride of 1 x 1), each followed by a max-pooling layer (pool size of 3 x 3 with a stride of 2 x 2), for facial feature extraction. On top, five independent convolutional branches compose the ensemble. Each branch was trained to learn different higher-level representations from earlier layers in order to increase diversity in the ensemble, and it consists of one convolutional layer (filter size of 3 x 3 with a stride of 1 x 1), and two fully-connected layers with 100 and 2 neurons (i.e., arousal and valence outputs). As activation functions, we utilized ReLU for the convolutional layers and the hyperbolic tangent function for the fully-connected layers which give us a prediction between -1 to 1 for valence and arousal. Although this approach is able to be continually re-trained using unlabelled samples based on the ensemble predictions, we did not use this property in our experiment to avoid any side effects.

We trained the network on the AffectNet dataset \cite{mollahosseini2017affectnet} which is composed of a thousand images of facial expressions gathered from social media and annotated by humans based on discrete and dimensional representations of emotion. As pre-processing, we cropped the faces using the facial coordinates provided by the dataset, normalized the pixel values from 0 to 1, and re-scaled the cropped faces to 96 x 96 pixels in order to reduce the computational cost. Our results on emotion recognition for the dimensional representation are comparable with Mollahosseini et al. \cite{mollahosseini2017affectnet}, which is 0.38 RMSE for arousal, and 0.44 RMSE for valence compared to 0.41 and 0.37 achieved by them in the validation set. We could not evaluate our system on the test set, since it has not been released when we trained our system.

In the HRI experiment, NICO detects the closest face by applying the Viola and Jones algorithm \cite{viola2004jones} in an interval of three times per second. The averages of arousal and valence for the two most expressive faces among the last four facial expressions are stored. In our affective visual perception module, we decided to adopt arousal as a modulation factor for valence in order to cover a wide range of affective states. According to the circumplex model of affect by Russel \cite{russell1980circumplex}, for example, content and delighted have similar values on the valence scale but are far from each other on the arousal scale. Therefore, the use of arousal value to modulate valence might provide more conclusive evidence on how positively or negatively the auditory stimulus was experienced by a subject than utilizing valence only. In order to preserve the polarity of the perceived emotion, the arousal prediction is normalized between 0 and 2. Once a stimulus is presented, an emotional score value for the stimulus played by NICO is computed by multiplying the arousal and valence averages from half of the most expressive expressions. The computation of arousal and valence averages over multiple chunks with the most prominent facial expressions for estimating the emotional score value acts as an attention mechanism to filter out the excessive number of neutral expressions during the auditory stimulus.

\subsection{Affective Language Perception Module}
\begin{figure}[h]
	\centering
	\includegraphics[width=.49\textwidth]{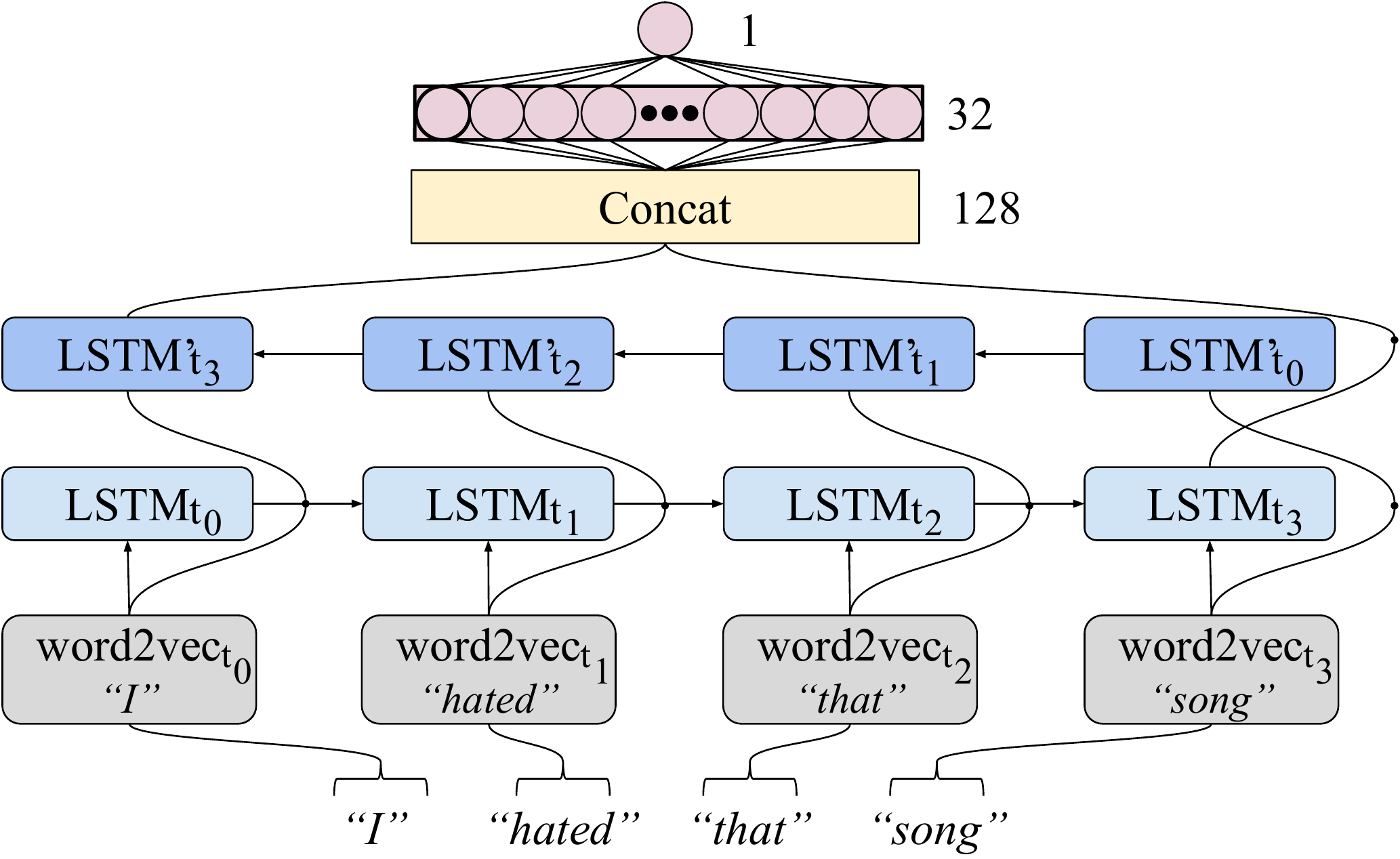}
	\caption{A bidirectional long short-term memory network for valence recognition from language features.}
	\label{fig:language}
\end{figure}

Subjects were asked to express, in natural language, how they felt about the song during the interaction. The text from spoken utterances acted as input for the affective language perception module. In order to recognize semantic polarity i.e. valence from language, a bidirectional long short-term memory network (BLSTM) with a hidden layer of size 64, an intermediate hidden layer of size 32, and a final sigmoidal output neuron was trained on the reader samples of the EmoBank dataset \cite{buchel2017emobank,buchel2017readers}. A hyperbolic tangent activation function was to the final hidden states of the BLSTM and ReLU was applied to the output of the fully connected layer. The model is visualised in Figure \ref{fig:language}. The EmoBank is a large text corpus of around ten thousand sentences that were annotated using a Valence-Arousal-Dominance scale through crowd-sourcing. The corpus was constructed of sentences from the Manually Annotated Sub-Corpus of the American National Corpus \cite{ide2008masc} and the corpus of the SemEval 2007 Affective Text task \cite{strapparava2007semeval}. Given that the EmoBank is labelled on a scale of zero to five we normalise these values from 0 to 1 to train the network. 

Sentences labelled with valence were converted to sequences of word embeddings through the commonly used Google News Vector pre-trained embedding\footnote{https://github.com/mmihaltz/word2vec-GoogleNews-vectors} which was trained using the word2vec algorithm \cite{mikolov2013efficient}. The sequences of word embeddings representing the sentences were used to train the BLSTM which learned how to interpret semantic polarity from the data. Only reader samples were selected, as these samples were deemed most suitable for recognising emotions from the language of others. The BLSTM was chosen as it has been shown to outperform unidirectional LSTMs on language tasks \cite{graves2005framewise} and it has the potential to capture additional semantic information because of its bidirectional nature, as it has access to both past and future words as context, whereas the LSTM is only able to look into the past. We achieved an RMSE of 0.07 on the normalised labels using the BLSTM trained on the reader samples of the EmoBank dataset, using 10\% of samples as test data.

The BLSTM was used to recognise user emotions when users were describing how they felt about a particular song. Subject utterances were converted to text via the top hypothesis provided by Google Speech\footnote{https://cloud.google.com/speech-to-text}. The sequences of words were then converted to word embeddings using the same pre-trained embedding as the one used to train the language model. The BLSTM then takes the sequence of word embeddings in order and in reverse order, concatenating the final hidden state outputs of both sequences before feeding them to the next fully connected layer. Finally, the fully connected layer is connected to a sigmoidal activation function, providing the valence classification. Once valence had been evaluated for a particular utterance, it was combined with the aforementioned vision evaluation to form the affective stimuli association for a particular song in order to determine NICO's next course of action. In order to align the prediction range of vision and language, the output valence for language is normalized to be between -1 and 1. We deemed that the strength of the semantic polarity was sufficient in order to evaluate the emotion, and thus, opted for the use of valence only for the affective language perception module.

\subsection{Affective Association Consolidation}
\begin{figure}[h]
	\centering
	\includegraphics[width=.495\textwidth]{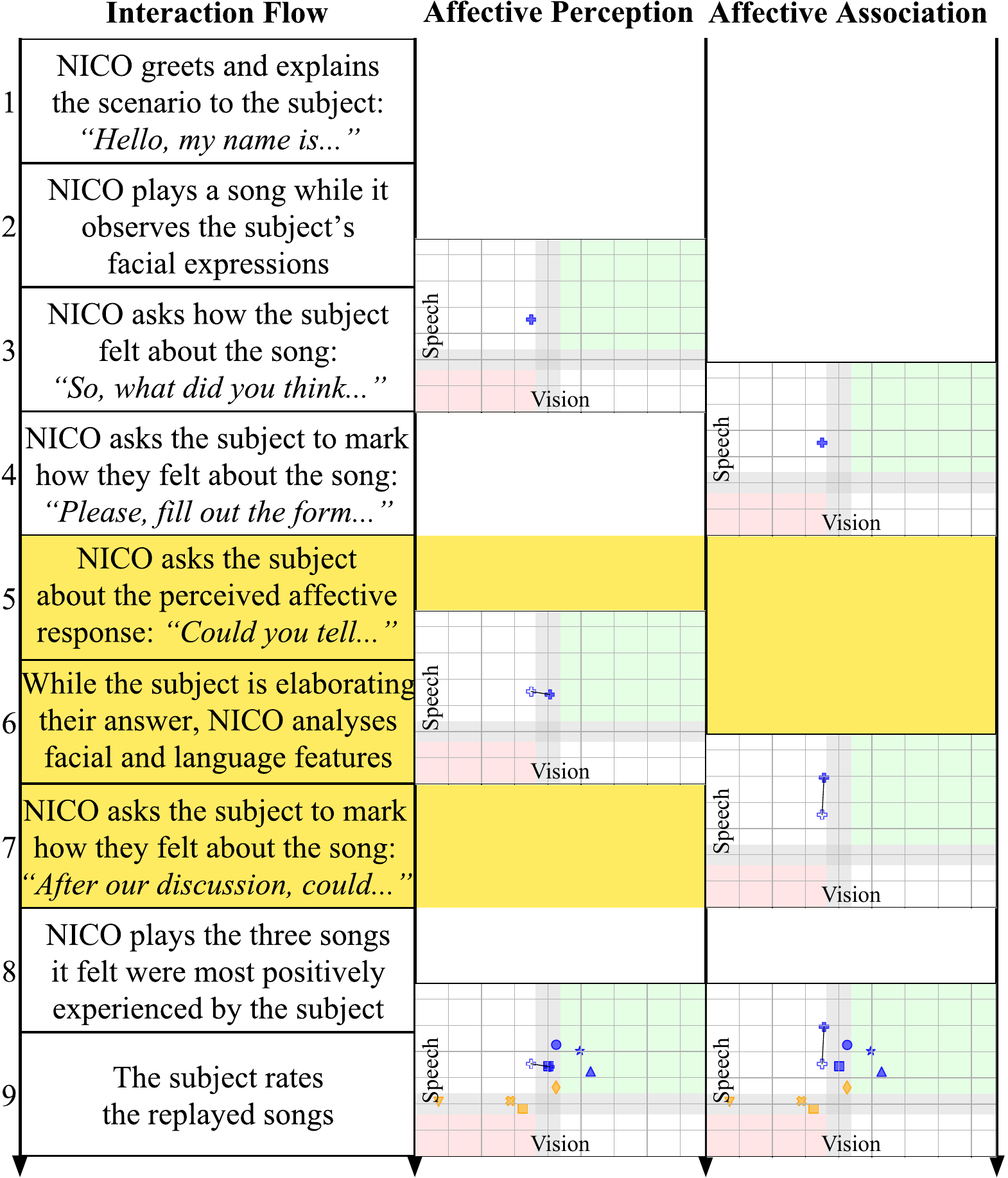}
	\caption{Illustration of the interaction flow for the conducted HRI experiment. The first column on the left shows each state during the interaction, whereas the second corresponds to NICO's affective perception of the subject after each recognition phase, and the last column on the right shows the affective information associated to a given stimulus. The yellow region represents a set of states after NICO finds a mismatching condition.}
	\label{fig:interactionflow}
\end{figure}

\begin{figure}[h]
	\centering
	\includegraphics[width=.495\textwidth]{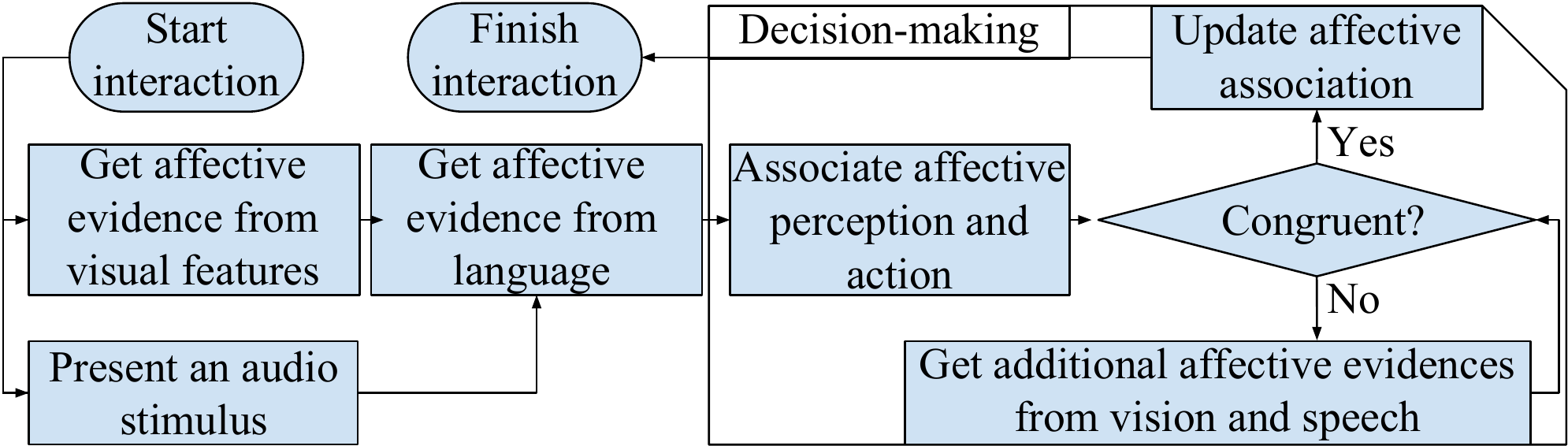}
	\caption{An overview of the operation cycle of our approach. After presenting an auditory stimulus and gathering affective information from vision and speech, the decision-making associates this information with a subject and verifies if more affective evidence is needed or not due to a possible mismatch between affective responses.}
	\label{fig:operationcycle}
\end{figure}

Once NICO has evaluated a subject's visual and language expressions, it has to consolidate these expressions into an affective association. The association process is visualised in Figure \ref{fig:interactionflow} and the entire decision-making process is visualised in Figure \ref{fig:operationcycle}. The initial association is mapped as the product of the valence and arousal of the visual expression against the valence of the language expression, denoted as the \textit{emotional score values}. This distributes the affective association into the two-dimensional affective association space, visualised in the first images in the second and third columns of Figure \ref{fig:interactionflow}. The association itself is represented as a blue icon in the figure.

Note that the affective association space can be divided into five different zones. The white top left and bottom right zones are incongruence zones. Associations that exist in these zones have conflicting evaluations of visual expressions and language. The top right green zone is a zone indicating coherent positive evaluations and the bottom left red zone indicates coherent negative evaluations. Finally, there is a neutral grey zone in the middle in the shape of a cross. Emotional score values within these grey areas indicate that one or both modalities evaluated the subject's expressions as neutral. Associations marked as neutral are considered to be coherent, as we lack the certainty to determine if an expression from a modality is positive or negative. The neutral region for both modalities has been set after a small pilot study to avoid mismatching conditions in low affective responses, which is 0.075 for vision and 0.04 for speech. 

If an association is placed in one of the white zones then it is considered to have an incoherent or mismatching evaluation. These will prompt NICO to further investigate the subject's evaluation of the perceived auditory stimulus through dialogue, as described by steps 5 to 7 in Figure \ref{fig:interactionflow}. In this case, NICO will analyse the subject's response to being asked to clarify their feelings toward the stimulus. It will analyse both the facial expressions and language used by the user synchronously, as opposed to the initial reaction which occurred asynchronously. 

Based on the new emotional score evaluations, NICO will adjust the believed association towards the stimuli. An example of how the perception may change can be seen in the second image in the second column of Figure \ref{fig:interactionflow}, represented by the shift from a hollow icon to a filled icon with a line between indicating the change between both modalities. However, NICO cannot simply replace the old perception with the new one, as this would invalidate the recorded initial emotional reaction to the stimulus. As such, it will use the new perceived response to modulate the original reaction. The affective association will only be adjusted when a final coherent expression is reached. 

In case of a mismatch, NICO will hypothesize that the most extreme modality (MEM) was correct, the MEM being either visual perception or language perception. If the new perceived expressions from the extended dialogue have the same MEM as the MEM from the original perception, then NICO's hypothesis is considered correct. The correct modality is boosted, and the other attenuated. We see an example of this behaviour for the second image in columns two and three of Figure \ref{fig:interactionflow}. In this case, language was the MEM, and when language again became the MEM in the extended dialogue, NICOs certainty in the language modality was significantly increased. We also see that NICO's belief in the vision modality does not necessarily increase just because coherent input was reached. Rather, only language was boosted as it was viewed as being more consistent and therefore more correct.

This is the primary difference between columns two and three in Figure \ref{fig:interactionflow}, as column two shows how the perceived emotions change with extended dialogue, whereas column three shows how NICO internally updates his understanding of the user's preference. Figure \ref{fig:interactionflow} also shows a practical example of this happening, where an incongruence is detected and NICO asks for clarification. NICO assumes that what the subject initially said was correct, rather than how they looked. During the clarification, the subject confirms NICO's suspicions and retains a positive explanation of the stimulus, albeit looking more positive. NICO will then take this as a confirmation that it was correct, update the users affective association toward that stimuli and continue the interaction. As an example of this entire procedure consider Figure \ref{fig:experimentalsetup}, where a person displays a positive facial expressions while giving negative language feedback. The person says that the song is terrible, NICO detects a mismatching condition and tries to gather additional evidence by continuing the dialogue until the subject responds cohesively. 

\section{EXPERIMENTS}
To collect data on how NICO performed when modelling affective data, experiments were conducted with 16 subjects to determine NICO's capabilities. The experiment is framed as a scenario where users are helping NICO to learn what music it should play, as illustrated in the accompanying video (\textit{https://goo.gl/XDCp3L}). The experiments are divided into two phases: learning and evaluation. In the first phase, NICO learns the most positively charged stimuli by presenting a stimulus and capturing the subject's emotional expressions.

Figure \ref{fig:interactionflow} illustrates the interaction flow for the conducted HRI experiment. During the learning phase, NICO presents nine different audio stimuli to a user. Each audio stimuli was, on average, 17 seconds long. Selected audio stimuli were a mixture between actual music represented in the figures by the colour blue; and a number of miscellaneous sound effects intended to confuse or irritate the user, represented by the colour orange. Even for music that users disliked they were expected to still be more positively inclined to these stimuli than to the miscellaneous sound effects.

Between each audio stimulus, NICO asked the user both to describe how they felt about the played ``song'' and then to rank how the song made them feel using a five-point scale. NICO associated the perceived affective response (middle column in Figure \ref{fig:interactionflow}) to its affective association space illustrated in the right column in Figure \ref{fig:interactionflow}. If NICO identified a mismatch in the emotion interpretation from perceived vision and language, then NICO initiated the additional dialogue wherein it attempts to achieve a coherent affective representation toward the audio stimulus by having the user elaborate on their feelings regarding the stimulus.

As indicated in Figure \ref{fig:interactionflow} by NICO perceiving a negative visual response in contrast to a positive description about a given song represented by the blue circles, NICO begins the additional dialogue by presenting a hypothesis based on the strongest perception modality. As an example, if the subject looked clearly positive but described the song as a negative experience, then the NICO leads with stating that it looked like the user enjoyed the experience before asking for elaboration. NICO will continue to analyse the subjects facial expressions and language while answering until both input modalities are aligned in their predictions or until a maximum limit of 5 extensions is reached (process described in the yellow region of Figure \ref{fig:interactionflow}). After gathering sufficient affective evidence about a given stimulus, NICO updates its affective association space with the novel evidence, and proceed the interaction by repeating the step 2. Note that, the states in the yellow region is only reached if a mismatching condition is satisfied.

After all songs are played, the HRI reaches to the step 8, where NICO plays the three songs it felt were most positively experienced by the subject. Finally, the subject rates how many of the replayed songs were in their top and bottom three out of all of the nine songs.

\section{RESULTS \& DISCUSSION}
\subsection{Modelling Stimuli}
\begin{figure}[h]
	\centering
	\includegraphics[width=.48\textwidth]{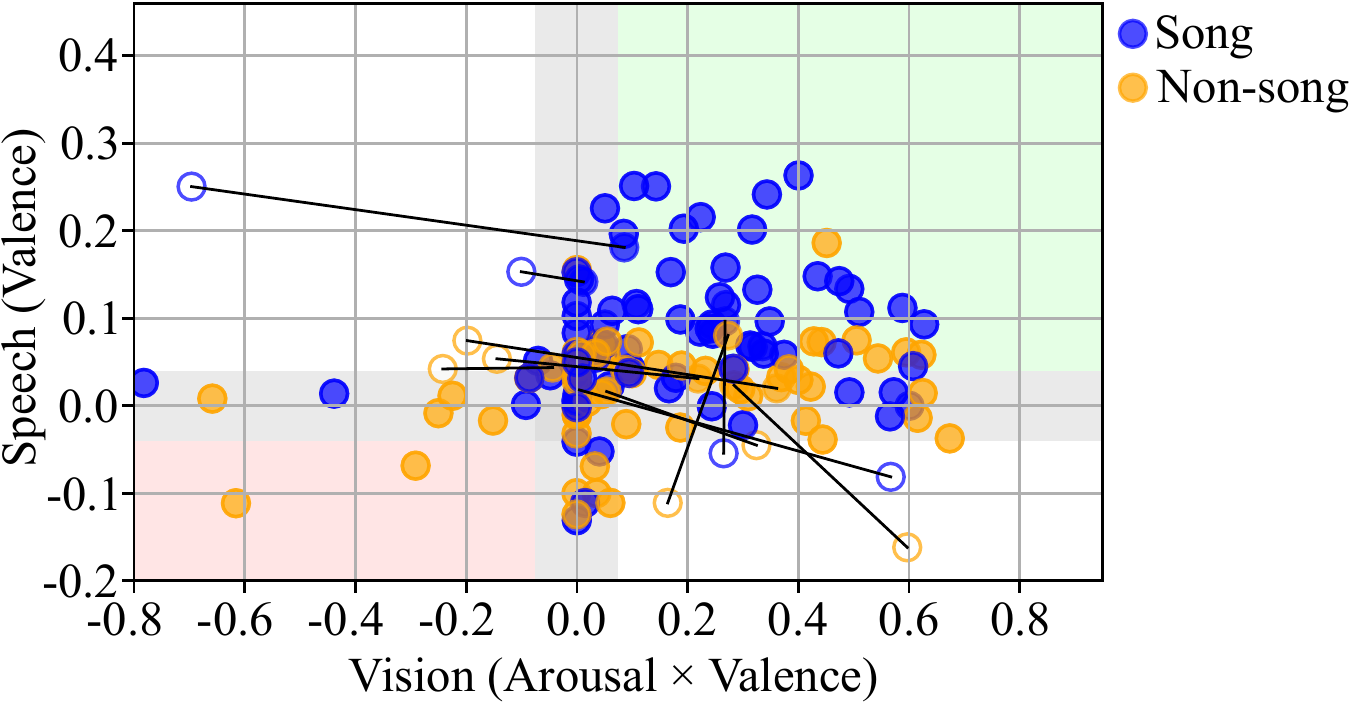}
	\caption{The clustering of how different subjects reacted to the same stimulus. Hollow circles with lines to full circles indicate the movement of a perceived auditory stimulus when the subject was asked to clarify their position after an extended dialogue.}
	\label{fig:PerceptionsComparison}
\end{figure}

\begin{figure}[h]
	\centering
	\includegraphics[width=.48\textwidth]{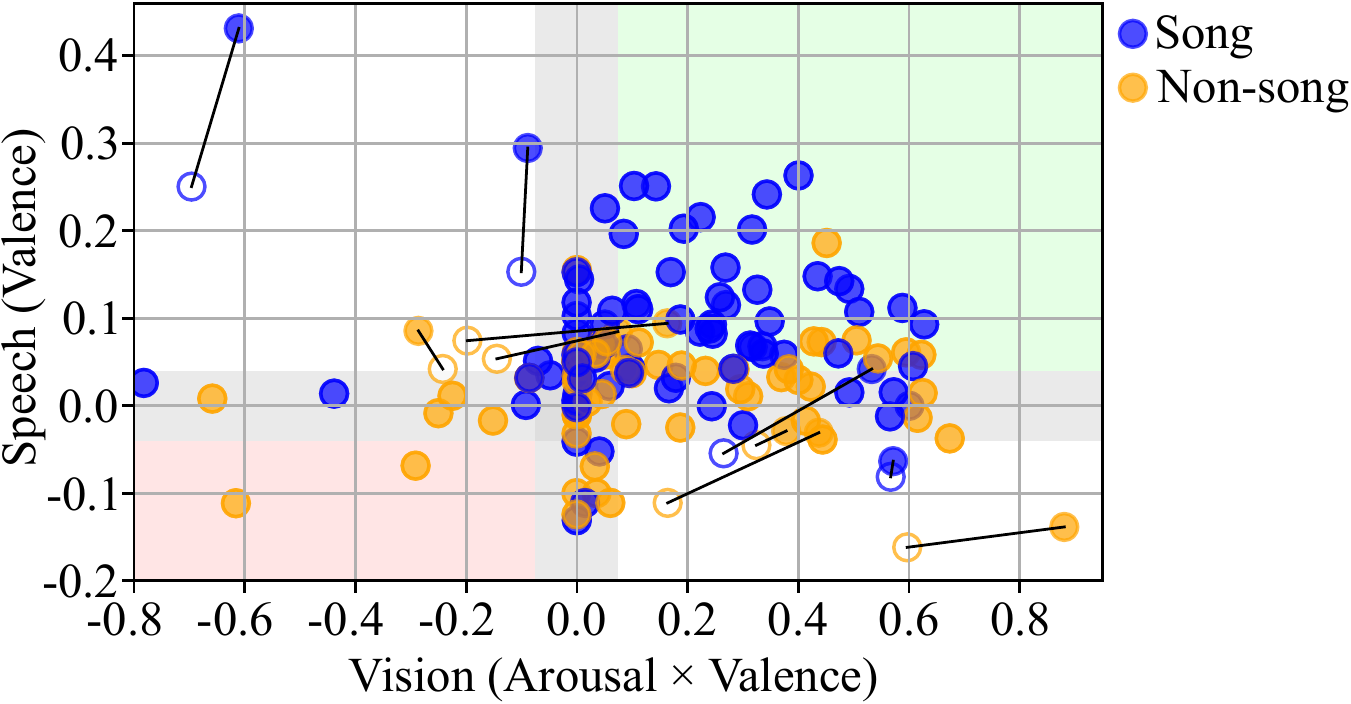}
	\caption{NICO's affective association space or affective memory. This image depicts where NICO thinks a subject's emotional state regarding a certain stimuli lies based on emotional expressions from speech and vision before (hollow circles) and after (full circles) the extended dialogue as illustrated by the adjustments.}
	\label{fig:MemoryComparison}
\end{figure}

\begin{figure*}[t]
    \centering
    \includegraphics[width=1\textwidth]{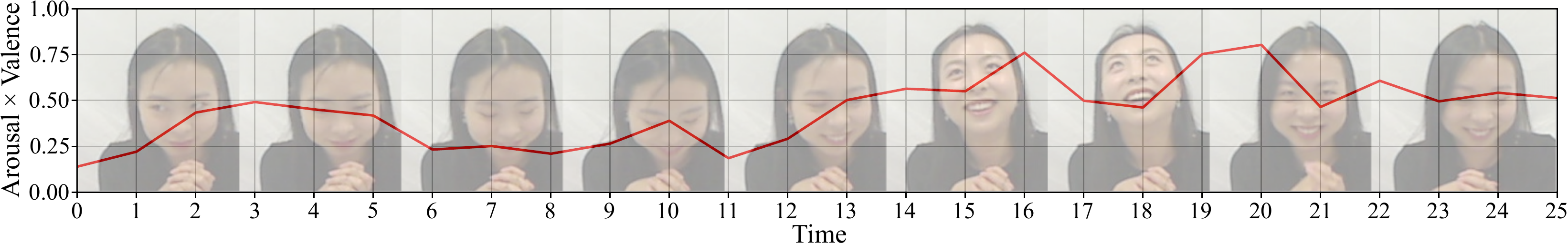}
    \caption{Visually perceived emotions of subject 05 for the fork sound over a period of 25 seconds.}
    \label{fig:subject5}
\end{figure*}

Despite the relatively low number of subjects in our experiment, trends can already be noticed in the clustering of data seen in Figure \ref{fig:PerceptionsComparison} and Figure \ref{fig:MemoryComparison}. The actual songs tend to be clustered far more frequently in the top right quadrant of the affective perception space and the affective association space. This indicates that NICO is able to determine songs from non-songs to a certain extent, by examining how subjects react to the stimuli. We initially expected more annoying sounds, such as chewing and a fork scraping, to be notably separated from actual songs, such as the classical Ode to Joy or a classic rock ballad. We noted that this is true when visualizing how different songs are distributed in the affective association space.

There are still a number of issues that remain, for example, very strong modalities are not investigated thoroughly enough. Figure \ref{fig:subject5} exemplifies this by showing a subject visually reacting very positively to a negative stimulus. The subject also subsequently described the audio stimulus in a manner that was negative but did not surpass the neutral boundary for speech. Nevertheless, this example shows the advantage of our system in comparison with its unimodal counterpart, where the latter might have led the NICO robot to wrongly associate a negative stimulus as positive if only vision is used.

\subsection{Modelling Individuals}
\begin{figure}[h]
	\centering
	\includegraphics[width=.48\textwidth]{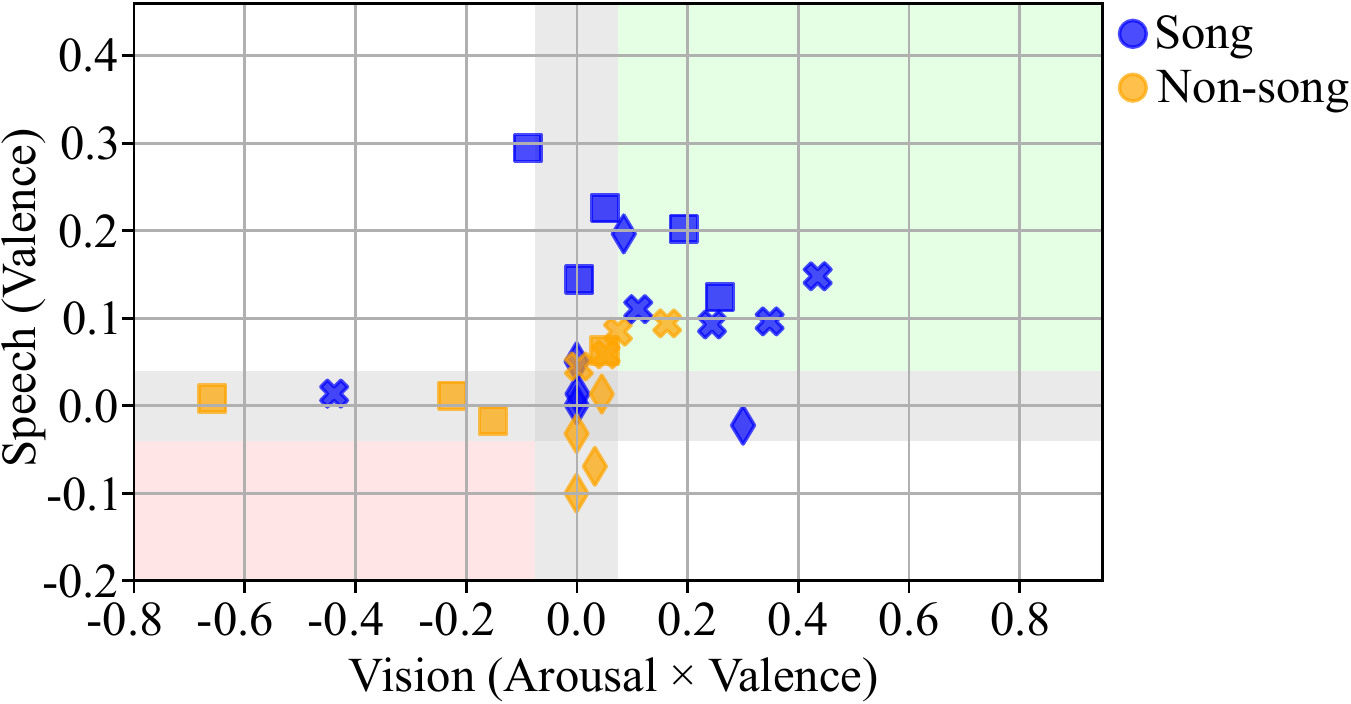}
	\caption{A visual comparison between three selected samples of the experiment participants distribution over the affective space, wherein each subject is represented by a different shape.}
	\label{fig:ComparisonSubjects}
\end{figure}

Similar to the distribution of songs, different subjects affective expression styles of subjects are also captured by NICO. In Figure \ref{fig:ComparisonSubjects} we see that the shapes representing different subjects distribute themselves in a particular manner unique to the subjects. We see that expressions from particular subjects tend to cluster in a specific manner, with a spread in song values being an indication of the expressiveness of users. As an example, the subject represented by the diamond had very minor fluctuations in regards to visual expressions and thus was not very expressive, whereas the square subject had varied expressions both verbally and visually. If users are more emotive in terms of language or facial expressions, we can see a larger variance in the affective space.

The existence of this spread in evaluations and how they are different for individuals provides valuable information on how to structure a system capable of effectively perceiving and resolving incoherence in perceived emotions. Currently, NICO uses a unified baseline for neutrality for all subjects. However, given that users may possess different expressive baselines, it may be prudent to have a dynamic and personalised baseline for every user.

\subsection{Resolving Incoherent Perception Through Dialogue}
The model attempts to form coherent affective associations by initiating extended dialogue when a mismatch of the perceived emotion from vision and language is detected. The practical effect of doing this can be seen in the shift of points depicted in Figure \ref{fig:PerceptionsComparison} and Figure \ref{fig:MemoryComparison}, wherein a hollow circle indicates a starting belief or perception, a full circle indicates an ending belief or perception, and the line between the two indicating the trajectory of the change for a specific shift. 

We see in Figure \ref{fig:PerceptionsComparison} that the influence of the extended dialogue on subjects subsequent expressions does shift the perceived expressions towards a coherent central line that can be drawn between the top-right and bottom-left quadrants of the figure. This indicates that subjects tend toward expressive coherence on subsequent elaborations on the stimulus.

The adjusted affective association when using the original perception as a baseline is depicted in Figure \ref{fig:MemoryComparison}. While it is possible for NICO to conclude coherent associations in some cases, the final association does not always reach coherent space. This is because NICO currently continues until coherent input is reached and not until NICO achieves a coherent association state for a stimulus. We found that out of all the samples we collected NICO never felt the need to extend the dialogue beyond a single additional interaction to collect additional information when resolving incoherence. 

\begin{figure*}[h]
    \centering
    \includegraphics[width=1\textwidth]{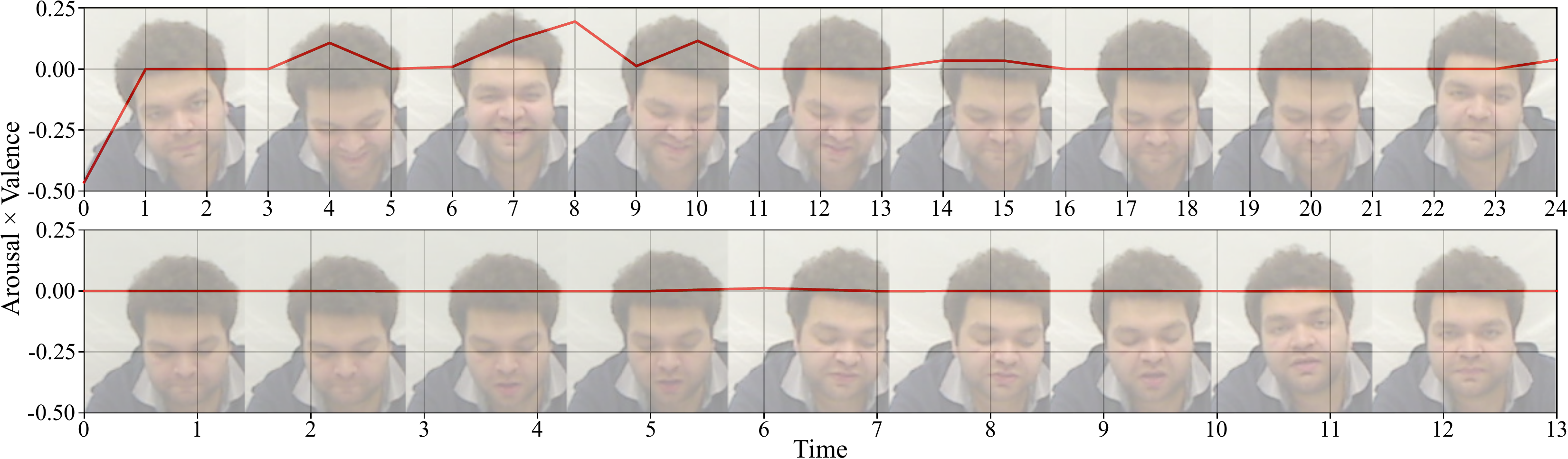}
    \caption{Visually perceived emotions of subject 03 for the Queen sound. On the top row is the initial reaction of vision for the actual presentation of the stimulus. The second row depicts the facial expression during the elaboration, which is notably more neutral. }
    \label{fig:subject3}
\end{figure*}

When performing extended dialogue, we also found that users tended toward more neutral facial expressions while retaining verbose language answers. This is exemplified in Figure \ref{fig:subject3} and manifests as the visual modalities perceptual transitions towards neutral zones in Figure \ref{fig:PerceptionsComparison}.

\section{CONCLUSIONS AND FUTURE WORK}
We presented a system that allows a NICO robot to model the relation between affective expressions and a perceived auditory stimulus. This system is able to model how audio stimuli are generally perceived by multiple users and how specific users differ from each other in emotional expressiveness. Furthermore, it is able to act upon accumulated affective information in order to be able to adjust its dialogue behaviour throughout the course of the interaction.

In future work, we aim to increase NICO's robustness to many of the issues identified during the experimental procedure, such as the metric for solved incongruence being based on coherent perception as opposed to a coherent belief. We also aim to improve NICO's flexibility regarding classifying emotions from different subjects by employing a dynamic neutral expression zone specific to individual users. Further structuring the NICO's interaction capabilities through intelligent dialogue will also be a focal point of future development, allowing NICO to make informed decisions in an interactive manner.

\section*{ACKNOWLEDGEMENTS}
This work has received funding from the European Union under the SOCRATES project (No. 721619), and the German Research Foundation under the CML project (TRR 169). We would like to thank Erik Strahl for his valuable contributions that improved the quality of this paper.

\bibliographystyle{IEEEtran}
\bibliography{refs}
\end{document}